\RequirePackage{fix-cm}
\documentclass[twocolumn]{svjour3}          
\smartqed  
\usepackage{graphicx}
\usepackage{cite}
\usepackage{amsmath,amssymb,amsfonts}
\usepackage{algorithmic}
\usepackage{array}
\usepackage{epsfig}
\usepackage{textcomp}
\usepackage{amssymb}
\usepackage{amsmath}
\usepackage{xcolor}
\usepackage{url}
\usepackage{pifont}
\usepackage{multirow}
\usepackage{booktabs}
\usepackage{lscape}
\usepackage{comment}
\usepackage{rotating}
\usepackage{rotating}

\begin{document}

\title{Detecting Autism Spectrum Disorder using Machine Learning
}
\subtitle{An Experimental Analysis on Toddler, Child, Adolescent and Adult Datasets}


\author{Md Delowar Hossain         \and
        Muhammad Ashad Kabir  \and
        Adnan Anwar \and
        Md Zahidul Islam
}


\institute{M D Hossain \at
              School of Computing and Mathematics \\
              Charles Sturt University\\
              \email{delowar.eee@gmail.com}           
           \and
           M A Kabir \at
             School of Computing and Mathematics \\
              Charles Sturt University\\
              NSW, Australia\\
              \email{akabir@csu.edu.au}
            \and
           A Anwar \at
             School of Information Technology \\
              Centre for Cyber Security Research \& Innovation (CSRI)\\
              Deakin University, Waurn Ponds, Australia\\
              \email{adnan.anwar@deakin.edu.au}
              \and
              M Z Islam \at
              School of Computing and Mathematics\\
              Charles Sturt University\\
              \email{zislam@csu.edu.au}
}

\date{Received: date / Accepted: date}

\maketitle

\begin{abstract}
Autism Spectrum Disorder (ASD), which is a neuro development disorder, is often accompanied by sensory issues such an over sensitivity or under sensitivity to sounds and smells or touch. Although its main cause is genetics in nature, early detection and treatment can help to improve the conditions. In recent years, machine learning based intelligent diagnosis has been evolved to complement the traditional clinical methods which can be time consuming and expensive. The focus of this paper is to find out the most significant traits and automate the diagnosis process using available classification techniques for improved diagnosis purpose. We have analyzed ASD datasets of Toddler, Child, Adolescent and Adult. We determine the best performing classifier for these binary datasets using the evaluation metrics recall, precision, F-measures and classification errors. Our finding shows that Sequential minimal optimization (SMO) based Support Vector Machines (SVM) classifier outperforms all other benchmark machine learning algorithms in terms of accuracy during the detection of ASD cases and produces less classification errors compared to other algorithms. Also, we find that Relief Attributes algorithm is the best to identify the most significant attributes in ASD datasets.


\end{abstract}

\section{Introduction} \label{sec:intro}
Autism spectrum disorder (ASD), is a neurological developmental disorder. It affects how people communicate and interact with others, as well as how they behave and learn \cite{Thabtah2019Regree}. 
The symptoms and signs appear when a child is very young. It is a lifelong condition and cannot be cured. A study conducted by Wiggin et al. found that 33\% of children with difficulties other than ASD have some ASD symptoms while not meeting the full classification criteria \cite{Wiggins2015a}.
ASD has a significant economic impact both due to the increase in the number of ASD cases worldwide, and the time and costs involved in diagnosing a patient. Early detection of ASD can help both patient and healthcare service providers by prescribing proper therapy and/or medication needed and thereby reducing the long-term costs associated with delayed diagnosis. On the other hand the traditional clinical methods such as Autism Diagnostic Interview Revised (ADI-R) and Autism Diagnostic Observation Schedule Revised (ADOS-R), are time consuming and cumbersome \cite{Lord1994, LeCouteur1995}.
The children who are too young and has delayed speech issue roughly score 25\% of the total ADI-R items because the verbal sections can’t be answered accurately for the patient. Besides, performing interview with a caregiver by a trained examiner takes 90-150 minutes. As a result, it is time-consuming and causes non-random missing data in summary score. On the other hand, the detection of ASD by ADOS-R depends on measurements of the scoring based on the answers provided. Moreover, one of the major disadvantages of this is the tendency to over classify children who have other clinical disorders\cite{Chawla2009}. So, the healthcare professionals are in urgent need of time efficient, easy and accessible ASD screening methods that can accurately detect whether a patient with a certain measured characteristic has ASD and inform individuals whether they should pursue a formal clinical diagnosis. Presently, the available datasets are few and associated with clinical diagnosis which is mostly genetic in nature, e.g., AGRE \cite{Geschwind}, National Database of Autism Research (NDAR)\cite{Pratap2014} and Boston Autism Consortium (AC) \cite{Fischbach}. Now a days, machine learning is applied to detect various diseases, e.g., depression~\cite{islam2018depression}, ASD~\cite{Pratap2014, Allison2012, Duda2016}, etc. The primary purpose of applying machine learning is to improve diagnosis accuracy and reduce diagnosis time of a case in order to provide quicker access to health care services. Since the diagnosis process of a case involves coming up with the right class (ASD, No-ASD) based on the input case features, this process can be attributed as a predictive task in machine learning.    

The purpose of applying classification techniques is to obtain improved precision, recall and predictive accuracy on the results of the feature selection methods. Furthermore, a comparison of the state-of-the-art classifiers has been performed considering learning errors and F-measures values. By classifying the ASD datasets and performing feature and predictive analyses, the below contributions have been achieved:

\begin{enumerate}
    \item We analyze the features of Toddler, Child, Adolescent and Adult ASD datasets, and present a correlation between the demographic feature and ASD cases.
    \item We explore benchmark feature selection methods and identify the most significant ones to classify ASD cases. Our analysis shows that appropriate feature selection can significantly improve the ASD classification performance.
    \item We compare the state-of-the-art classification methods and identify the best performing classifier that is suitable for all four ASD datasets.

\end{enumerate}

This paper is structured as follows: Section 2 gives us the literature review of autism screening methods, section 3 presents the analysis methodology. The description of dataset and exploratory analysis using the ASD dataset is presented in Section 4. The performance comparison using benchmark algorithms is presented in Section 5. In that section, we also highlighted the feature extraction results and summarized the performance evaluation with extensive experimental results. Finally, the paper concludes with some brief remarks in Section 6. 

\section{Related Work}\label{sec:relatedwork}
A number of researches have adopted machine learning techniques to improve the diagnosis process of ASD (e.g.,~\cite{Pratap2014, Allison2012, Duda2016}). The primary motivation behind the utilization of machine learning models on ASD is to reduce the detection time that will enable quicker access to health care services, improvement in diagnoses accuracy\cite{Thabtah2017}.  We can categorize the ASD detection techniques in two categories- Video clip based study and AQ- based study.

\subsection{Video clip-based study}
Tariq et al.\cite{Parikh2019}, have hypothesized that the use of machine learning analysis on home video can speed up the diagnosis time without compromising accuracy. Authors have analyzed item-level records from 2 standard diagnostic instruments to construct machine learning classifiers optimized for sparsity, interpretability, and accuracy. Authors have considered 8 machine learning models to apply on 162 two-minute home videos of children with and without autism diagnosis. Besides, a mobile web portal has been created for video raters to assess 30 behavioral features (e.g., eye contact, social smile) that are used by 8 independent machine learning models for identifying ASD. The result shows that 94\% accuracy is achieved for each case using cross-validation testing and subsequent independent validation from previous work. However, this method is also time consuming since the video needs to be recorded and assessed for the rating based on 30 questions. Whereas we adopt a method that only uses mobile app from where users can easily select the appropriate answers of the ten ASD questions. Besides, improved analysis based on reduced number of attributes number can significantly improve the performance of the ASD detection.\\
Andrea et al. \cite{Zunino2018} analyze video gesture for detecting ASD. Authors have devised experiment by recording video of patient and healthy children performing simple gesture of grasping a bottle. By only processing the video clips depicting the grasping action using a recurrent deep neural network, they are able to discriminate with a good accuracy considering binary classification- ASD and no-ASD. Authors have used neural network based model such as the Long Short-Term Memory (LSTM) network. In that work, authors followed a common procedure where each video is split in 15-frame clips and passed through the entire model that outputs a binary vector containing, for each frame, the probabilities for ASD or No-ASD. Each clip is considered independent during the training. The test accuracy for each subject is computed by averaging the scored probabilities over all the frames of the videos. But the model decreases its effectiveness after a threshold of 0.9. The results support the hypothesis that feature tagging of home videos for machine learning classification of autism can yield accurate outcomes in short time frames, using mobile devices.

\subsection{AQ based study}
Autistic-Spectrum Quotient (AQ) is a screening method developed by Baron-Cohen et al.\cite{Baron-Cohen2001}. Later Allison et al.\cite{Allison2012} propose the AQ 10-adult and AQ 10-child, shortened versions of the original AQ. This questionnaire based attempt is said to increase the efficiency of the ASD screening.

In recent times, Kanad et al. have analyzed the Adult Autism screening data set using supervised machine learning techniques such as Decision Trees, Random Forest, Support Vector Machines (SVM), k-Nearest Neighbors(kNN), Naïve Bayes, Logistic Regression, Linear Discriminant Analysis (LDA) and Multi-Layer Perceptron (MLP) \cite{kbasu}. Authors conclude that SVM classification technique works very well for Adult ASD dataset which gives the highest accuracy, recall and F-measure among all other benchmark techniques. Later Brian McNamara et al.\cite{BrainMc} also classify the same dataset by applying Decision Tree and Random Forest classifiers considering improved data pre-processing where authors remove some records for missing categorical instances and remove less significant variables before applying the classifiers. Authors compare between these two classifiers’ results and found that Random Forest gives more accuracy and predictability on Adult Autism Screening. Later M.D Hossain et. al. conducted the research on child ASD dataset \cite{UCIFadi:2017}, following the same methodology and applied 27 supervised classifiers to detect child ASD\cite{hossain2019detecting} cases. They have also found that the Support vector classifier- SMO peforms best to detect child ASD cases. Beside this, they were also able to identify dominant features in detecting child autism. Now, in our current study, we have conducted research on ASD datasets \cite{Fadi:2020} taking the consideration of four aged groups such as Toddler, Child, Adolescent and Adult. In order to obtain a clearer understanding towards ASD detection. Authors have developed Rules based Machine Learning algorith \cite{thabtah2020new} and applied on Child, Adolescent and Adult datasets. Based on the available ASD dataset, our goal is to analyze all four datasets (e.g., 1. Toddler, 2. Child, 3. Adolescent and 4. Adults) by applying 25 well known supervised classification techniques in order to classify them in one of two categories: “Patient having ASD” or patient doesn’t have ASD. One important aspect will be selecting critical features in order to improve the detection accuracy and efficiency and select the best performing algorithm.

\section{Methodology}

\begin{figure*}[htbp]
    \centering
    \includegraphics[width=1\textwidth]{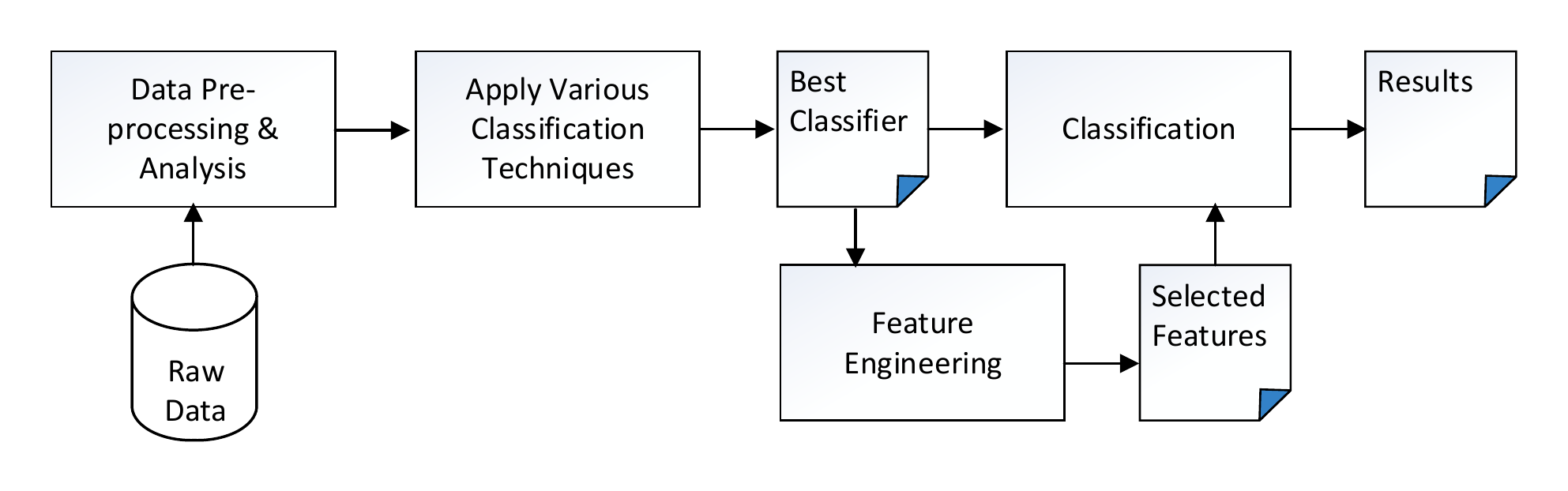}
    \caption{Classification flow chart for autism screening}
    \label{fig:flow}
\end{figure*}

Our objective is to find out the most significant attributes and automate the process using classification techniques and finally identify the best performing algorithm. We have analysed ASD datasets of Toddler, Child, Adolescent and Adults towards determining the best performing classifier for these binary datasets considering recall, precision, F-measures and classification errors.We pre-process our dataset by removing the attributes that have missing values and also those that do not offer any benefit during the analyses. Fig \ref{fig:flow} shows an overview of our methodology for detecting ASD cases.  In this framework, output label (ASD or no -ASD) will be assessed by the best classifier for the individual ASD screaming process. Besides, we also determine the attribute counts and the corresponding F-measure values to identify the most significant attributes. During our analysis, we have adopted a systematic approach, briefly described below. 

\begin{itemize}
\item \textbf{Step 1: Data prepossessing \& analysis:} 
In this step, we performed a deatiled data-preprocessing before the going in to the detailed analysis and classification. The collected datasets from \cite{Fadi:2020} need to be cleaned as the datasets contain few records with missing values. So, those records have been removed and was not considered for classification. Besides, we also discard attributes that do not offer any benefit. In this phase, the characterization of the datasets is also evaluated. Section 4 presents the dataset details and explore the details of the datasets, e.g., we investigate the demographic impact  such  as  Jaundice,  family  ASD  and  Ethnicity  for the ASD datasets. Some critical steps related to the variable reduction is presented in Section 5.1.\\

\item \textbf{Step 2: Applying classification techniques to identify best classifier:}
It this step, we applied various classification techniques. At first, we build the machine learning training  model. We use k-fold cross validation to train our dataset for evaluation purpose. In this study we have applied 25 classification techniques to all our four datasets. We present this information in Section 5, specifically, Section 5.2 summarizes the detailed evaluation matrices and Section 5.4 highlights the error matrices.\\

\item \textbf{Step 3: Feature engineering:}
We determine the best classifier and then we  classify the cases. Beside these, we do feature engineering and run the classification using selected features. We continue this process to determine the least feature with highest classification recall and accuracy. The complete procedure is presented in Section 6. While Section 6.1 describes different feature selection techniques and feature ranking, Section 6.2 identify the most significant feature. \\

\item \textbf{Step 4: Classification \& results comparison:}
In this stage the classification task is performed to identify whether an instance in the datset has ASD or No ASD. The results from ASD-yes cases emphasis that the person needs to undergo further medical advice and necessary treatments. Therefore, higher accuracy is important to reduce the false positive and overall improvement of the outcome. Section 7 presents a comparative analysis of the classification results. Considering four different datsets, we show that our proposed method outperform existing benchmark techniques.

\end{itemize}

\begin{figure}
\centering
    \includegraphics[width=1\columnwidth]{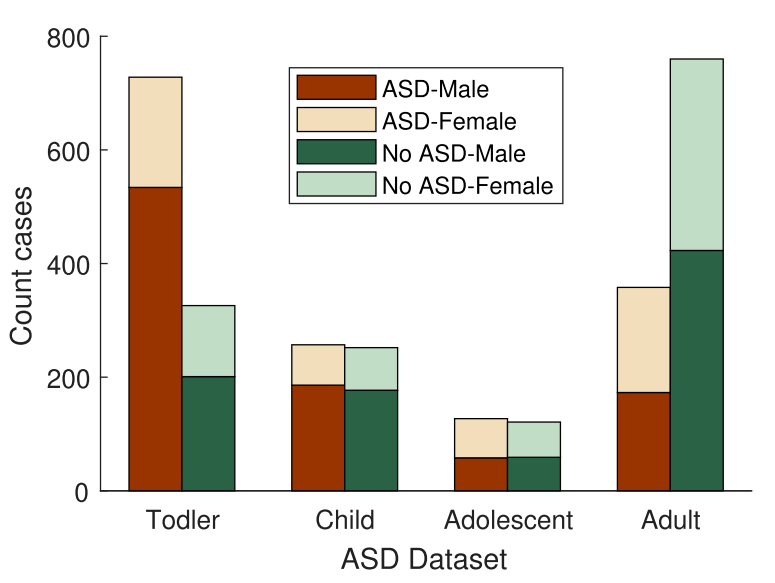}
    \caption{Gender-wise distribution of ASD cases in four datasets (Todler, Child, Adolescent and Adult)}
    \label{fig:dist}
\end{figure}

\begin{figure}[htbp]
\centering
    \includegraphics[width=1\columnwidth]{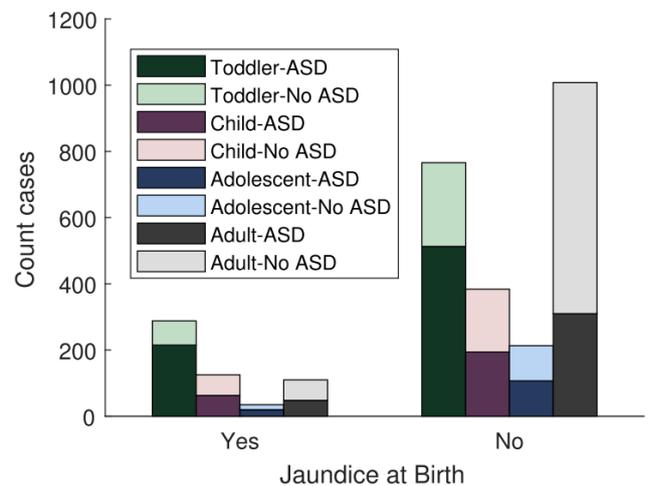}
    \caption{Impact of Jaundice at Birth on ASD}
    \label{fig:Jaund}
\end{figure}

\section{Data Preprocessing \& Analysis}
\subsection{Dataset Description}
Our colleted ASD dataset~\cite{Fadi:2020} mainly consists of 23 attributes (except Toddler dataset where we consider 18 attributes). The attribute descriptions are given in Table~\ref{table:dataset}. All datasets have ten binary attributes representing the screening questions (A1 to A10) as well as the categorical variables such as gender, ethnicity, jaundice, Family\_ASD, Residence and ASD class. Datasets also have two numeric variables such as Age and Screen Score/Results. We find that some attributes are absent in Toddler ASD dataset such as : who completed the test (User), Why taken the screening, Used\_App\_Before, Country of Residence and Language Spoken.

We find that Child and Adolescent datasets have almost similar questionnaires whereas Toddler and Adult questionnaires show variations. Each question is associated with multiple answers that can be easily selected in a mobile environment using a smartphone. We have presented various questionnaires of ASD dataset according to the sequence for Child, Adolescent, Adult and Toddler (please see the Description column for details) in Table \ref{table1lbl}. The class type is assigned during the process of data collection by answering the AQ-10 (A1 to A10) questions. So, the class value “No” is assigned when the final score based on AQ-10 methods scores less than or equal to 7. Otherwise, it is assigned a “Yes” which indicates that the individual does have the ASD. But for Toddler dataset the cut-off score is less than or equal to 4. So, in this case if the total score $\geq$ 4, it is considered that the subject has ASD. Figure \ref{fig:dist} shows the class distribution of the considered dataset. Here, we observe that the child and adolescent dataset are somehow balanced but Toddler and Adult datasets are not balanced considering the total number of ASD cases and gender distribution.

\begin{table*}
\centering
\caption{ASD Dataset Description}
\label{table:dataset}
\begin{tabular}{p{3cm}p{3cm}p{9cm}}
\hline
\textbf{Attribute}~                        & \textbf{Type}                   & \textbf{Description}                                                                                                                                                                                                                                                                                                                                                                                                                                                                                            \\ 
\hline
\hline
Age                               & Number                 & Age in months/years                                                                                                                                                                                                                                                                                                                                                                                                                                                                                    \\ 
\hline
Gender                            & String                 & Male or Female                                                                                                                                                                                                                                                                                                                                                                                                                                                                                         \\ 
\hline
Ethnicity                         & String                 & List of common ethnicities in text format                                                                                                                                                                                                                                                                                                                                                                                                                                                              \\ 
\hline
Born with jaundice                & Boolean~ (yes
  or no) & Whether the case was born with jaundice                                                                                                                                                                                                                                                                                                                                                                                                                                                                \\ 
\hline
Family member with PDD            & Boolean~ (yes
  or no) & Whether any immediate family member has a PDD                                                                                                                                                                                                                                                                                                                                                                                                                                                          \\ 
\hline
Who is completing the test (User) & String                 & Parent, self, caregiver, medical staff,
  clinician,etc.                                                                                                                                                                                                                                                                                                                                                                                                                                               \\ 
\hline
Why taken the screening           & Meta                   & The person can write short reason for completing the
  task                                                                                                                                                                                                                                                                                                                                                                                                                                            \\ 
\hline
Used\_App\_Before                 & Boolean~ (yes
  or no) & This answer would be binary                                                                                                                                                                                                                                                                                                                                                                                                                                                                            \\ 
\hline
Language spoken                   & String                 & The user will give his/her native language
  information                                                                                                                                                                                                                                                                                                                                                                                                                                               \\ 
\hline
Country of residence              & String                 & List of countries in text format                                                                                                                                                                                                                                                                                                                                                                                                                                                                       \\ 
\hline
Used the screening app before     & Boolean (yes or no)    & Whether the user has used a screening app                                                                                                                                                                                                                                                                                                                                                                                                                                                              \\ 
\hline
Screening Method Type             & Integer(0,1,2,3)       & The type of Screening Methods choses based on age
  category (0=toddler, 1=Child, 2= Adolescent, 3= Adult)                                                                                                                                                                                                                                                                                                                                                                                             \\ 
\hline
Question 1 (A1)                    &
  Binary (0, 1)
  ~  & S/he often notices small sounds when others do not,
  (Child, Adolescent)
  S/he notices patterns in things all the time, (Adult)
  Does your child look at you when you call his/her
  name? (Toddler)                                                                                                                                                                                                                                                                                                \\ 
\hline
Question 2 (A2)                        & Binary (0, 1)          & S/he usually concentrates more on the whole picture,
  rather than the small detail, (child, Adolescent, Adults)
  How easy is it for you to get eye contact with your
  child? (Toddler)                                                                                                                                                                                                                                                                                                              \\ 
\hline
Question 3 (A3)                        & Binary (0, 1)          & In a social group, s/he can easily keep track of
  several different people’s conversations, (child, Adolescent)
  I find it easy to do more than one thing at once,
  (Adult)
  Does your child point to indicate that s/he wants
  something? (e.g. a toy that is out of reach) (Toddler)                                                                                                                                                                                                            \\ 
\hline
Question 4 (A4)                        & Binary (0, 1)          & S/he finds it easy to go back and forth between
  different activities, (child, Adolescent)
  If there is an interruption, s/he can switch back to
  what s/he was doing very quick, (Adult)
  Does your child point to share interest with you?
  (e.g. pointing at an interesting sight) (Toddler)                                                                                                                                                                                                   \\ 
\hline
Question 5 (A5)                        & Binary (0, 1)          & S/he doesn’t know how to keep a conversation going
  with his/her peers, (child, Adolescent)
  I find it easy to read between the lines when someone
  is talking to me, (Adult)
  Does your child pretend? (e.g. care for dolls, talk
  on a toy phone) (Toddler)                                                                                                                                                                                                                                     \\ 
\hline
Question 6 (A6)                    & Binary (0, 1)          & S/he is good at social chit-chat, (child, Adolescent)
  I know how to tell if someone listening to me is
  getting bored, (Adult)
  Does your child follow where you’re
  looking? (Toddler)~~~~~~~~~                                                                                                                                                                                                                                                                                                  \\ 
\hline
Question 7 (A7)                     & Binary (0, 1)          & When s/he is read a story, s/he finds it difficult to
  work out the character’s intentions or feelings, (Child)
  When s/he was younger, s/he used to enjoy playing
  games involving pretending with other children, (Adolescent)
  When I’m reading a story, I find it difficult to work
  out the characters’ intentions, (Adult)
  If you or someone else in the family is visibly
  upset, does your child show signs of wanting to comfort them? (e.g. stroking
  hair, hugging them (Toddler)  \\ 
\hline
Question 8 (A8)                     & Binary (0, 1)          & When s/he was in preschool, s/he used to enjoy
  playing games involving pretending with other children, (Child)
  S/he finds it difficult to imagine what it would be
  like to be someone else, (Adolescent)
  I like to collect information about categories of things
  (e.g. types of car, types of bird, types of train, types of plant etc),
  (Adult)
  Would you describe your child’s first words as:
  (Toddler)                                                                            \\ 
\hline
Question 9 (A9)                    & Binary (0, 1)          & S/he finds it easy to work out what someone is
  thinking or feeling just by looking at their face, (Child)
  S/he finds social situations easy, (Adolescent)
  I find it easy to work out what someone is thinking
  or feeling just by looking at their face, (Adult)
  Does your child use simple gestures? (e.g. wave
  goodbye) (Toddler)                                                                                                                                                         \\ 
\hline
Question 10 (A10)                      & Binary (0, 1)          & S/he finds it hard to make new friends, (Child,
  Adolescent)
  I find it difficult to work out people’s intentions,
  (Adult)
  Does your child stare at nothing with no apparent
  purpose? (Toddler)
  ~                                                                                                                                                                                                                                                                                            \\ 
\hline
Screening Score 
  ~
  ~          & Integer (0 to 10)             & The final score obtained based on the scoring
  algorithm of the screening method used. This is computed in an automated
  manner                                                                                                                                                                                                                                                                                                                                                                      \\ 
\hline
Class                             & Binary (0, 1)          & Whether the objective has identified as ASD or not                                                                                                                                                                                                                                                                                                                                                                                                                                                     \\
\hline

\hline
\end{tabular}
\end{table*}

\subsection{Data Pre-processing}
In order to simplify our model and to get accurate results, we need to clean up our datasets by removing the fields corresponding to the missing values. Next, we aim to preprocess the dataset by reducing the less important attributes.

We have observed that in our dataset there are couples of attributes which are less significant and do not offer any benefit to our analysis. These are listed below:
\begin{itemize}
    \item Case
    \item Used App Before
    \item User (who completed the screening
    \item Language
    \item Why taken the screening 
    \item Age Description
    \item Screening Type
    \item Score
\end{itemize}

We find from the observed data, based on the score column, a result value of 7 or higher will always be classified as ASD\_Class= YES for Child, Adolescent and Adult ASD datasets. Similarly, score value 4 or higher will always be classified as ASD\_Class= YES for Toddler dataset. So, including these attributes imply that the classification algorithm already has the outcome of Target variable. For this reason those attributes are removed during analysis. 
Finally, we get 17 attributes to work with for Child, Adolescent and Adult dataset and 16 attributes (“Residence” attribute is missing) for Toddler datasets.

In the next subsections, we investigate the impact of Jaundice, family ASD and Ethnicity on ASD cases for all four ASD datasets.  
\begin{table*}[t]
    \centering
    \caption{Level of significance using pearson correlation (2-tailed) test}
    \label{tab:correlation-result}
    \begin{tabular}{ccccc}
    \hline
    \multicolumn{2}{c}{Variable} & p-value & Correlation coefficient (r) &  Strength of correlation\\
    \hline
    \hline
    \multirow{4}{*}{Jaundice at birth} & Toddler & .016 & .074 & negligible ($r\leq.25$)\\
    & Child & .999 &.000 & no significant ($p\geq.05$)\\
    & Adolescent & .451 & .048 & no significant ($p\geq.05$)\\
    & Adult & .007 & .081 & negligible ($r\leq.25$)\\
    \hline
     \multirow{4}{*}{{Family ASD}} & Toddler & .661 & -.014 & no significant ($p\geq.05$)\\
     & Child & .752 & .014 & no significant ($p\geq.05$)\\
     & Adolescent & .860 & -.011 & no significant ($p\geq.05$)\\
     & Adult & .000 & .151 & negligible ($r\leq.25$)\\
     \hline
     \multirow{4}{*}{{Ethnicity}} & Toddler & .207 &.039 & no significant ($p\geq.05$)\\
     & Child & .490 & -.031 & no significant ($p\geq.05$)\\
     & Adolescent & .001 & .217 &  negligible ($r\leq.25$)\\
     & Adult & .000 & -.174 & negligible ($r\geq-.25$)\\
     
    \hline
    
    \hline
    \end{tabular}
    
\end{table*}

\subsection{Jaundice at Birth}
The stacked bars in Fig \ref{fig:Jaund} show how the ASD cases are divided into jaundice at birth cases for all four datasets.
Our correlation analysis results in Table~\ref{tab:correlation-result} as well as the Fig.~\ref{fig:Jaund} show that there is no significant correlation between Jaundice at birth and ASD for Child and Adolescent, in particular. Whereas the result for Toddler and Adult dataset have shown some significance (as $p<.05$) but the correlation coefficient founds very negligible (as $r<.1$). 

\subsection{Impact of Family ASD}
\begin{figure}
\centering
    \includegraphics[width=1\columnwidth]{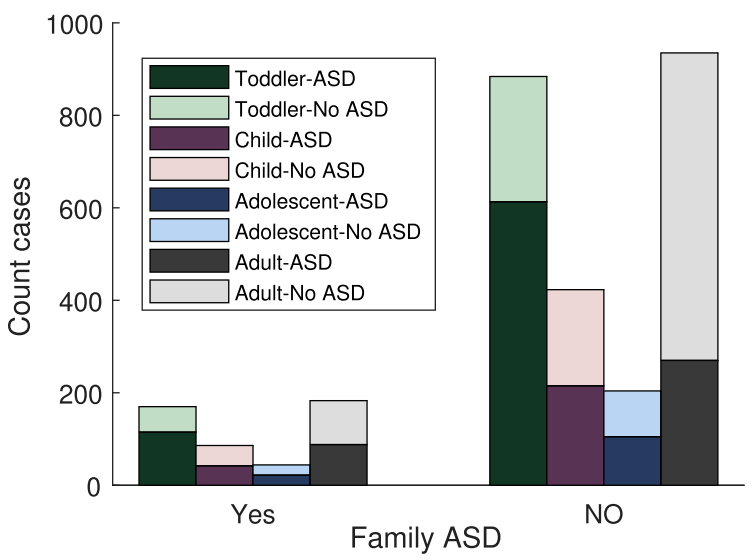}
    \caption{Impact of Family ASD on their offspring ASD}
    \label{fig:Family_ASD}
\end{figure}

The bar chart in Fig.~\ref{fig:Family_ASD} shows how the ASD cases are divided into Family ASD cases for all four datasets. The bars in the Fig.~\ref{fig:Family_ASD} and the correlation analysis results in Table.~\ref{tab:correlation-result} show that there is no significant correlation between Family ASD cases and their children ASD cases for Toddler, Child and Adolescent. Whereas we got a significant p-value ($p<.001$) for Adult dataset but the resulted r-value ($r<.2$) makes the strength of the correlation negligible.


\subsection{Impact of Ethnicity}

\begin{figure*}[htbp]
    \centering
    \includegraphics[width=1\textwidth]{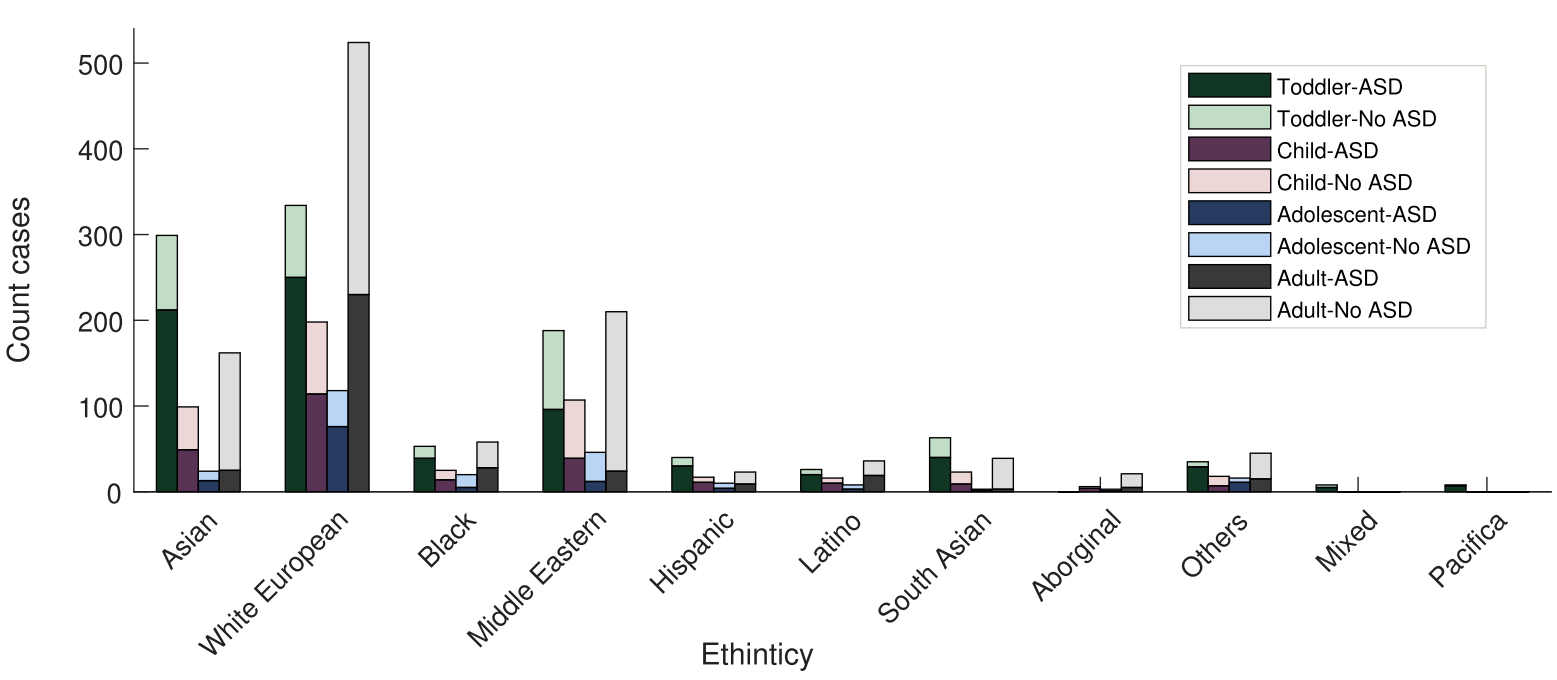}
    \caption{Ethnicity in ASD datasets}
    \label{fig:etch}
\end{figure*}

Fig.~\ref{fig:etch} shows how the ASD cases are divided into various ethnicity for all four datasets. The bars in the Fig.~\ref{fig:etch} and the correlation analysis results in Table~\ref{tab:correlation-result} show that there is no significant correlation between ethnicity and ASD cases, in particular for Toddler and Child. Whereas we got a significant p-value ($p<.001$) for Adolescent and Adult datasets but the resulted r-value (between 0 to $\pm.25$) makes the strength of the correlation negligible. 

\section{Analysis of Classification Techniques}
We apply 25 benchmark machine learning algorithms to determine the best classification model that supports our observed data. No single model is universally perfect for all types of classification problems. So, in order to determine the best model, we need to pre-process our data. For the evaluation purpose, we compare their performance by using confusion matrix.

\subsection{Evaluation matrix}
For a given dataset and a predictive model, every data point will lie on one of the below four categories.

True Positive (TP) -- The individual having ASD and was correctly predicted as having ASD. True Negative (TN) -- The individual not having ASD and was correctly predicted as not having ASD. False Positive (FP) -- The individual not having ASD, was incorrectly predicted as having ASD. False Negative (FN) -- The individual having ASD was incorrectly predicted as not having ASD.

Above four categorized data form a matrix which is called Confusion matrix (see Table~\ref{tab:Confusion Matrix}. This allows to visualize the performance of a supervised machine learning.\\
\begin{table}[h]
\caption{Confusion Matrix}
\label{tab:Confusion Matrix}

\begin{tabular}{ p{1.3cm} | p{2.8cm}  p{3.2cm} }
\hline
Prdiction  & Observed with ASD & Observed with No ASD \\
\hline
\hline
ASD=No  & False Negative & True Negative\\

ASD=Yes & True Positive & False Positive\\                 
\hline  

\hline
\end{tabular}

\end{table}

Accuracy:  
It is the measures of correct predictions made by the classifier. Accuracy is the number of correctly identified predictions by total number of predictions:
\begin{equation}
    Accuracy=\frac{TP+TN}{TP+FP+FN+TN}
\end{equation}
                                     
Precision:  
It measures the accuracy of positive predictions. It is the ratio of true positive out of the total observed positive.
\begin{equation}
   Precision=\frac{TP}{TP+FP}   
\end{equation}
                                                      
Recall/Sensitivity: 
This is also called true positive rate. It is the proportion of samples that are genuinely positive by all positive results obtained during the test. 

\begin{equation}
    Recall=\frac{TP}{TP+FN}
\end{equation}
                                                      
F-Measure: 
The F-score (or F-measure) considers both the precision and the recall of the test to compute the score. The traditional or balanced F-score (F1 score) is the harmonic mean of the precision and Recall:
\begin{equation}
F-Measure =\frac{2\times Precision}{Precision+Recall}
\end{equation}

\subsection{Comparison of Classification Techniques}
\begin{figure*}[t]
    \centering
    \includegraphics[width=1\textwidth]{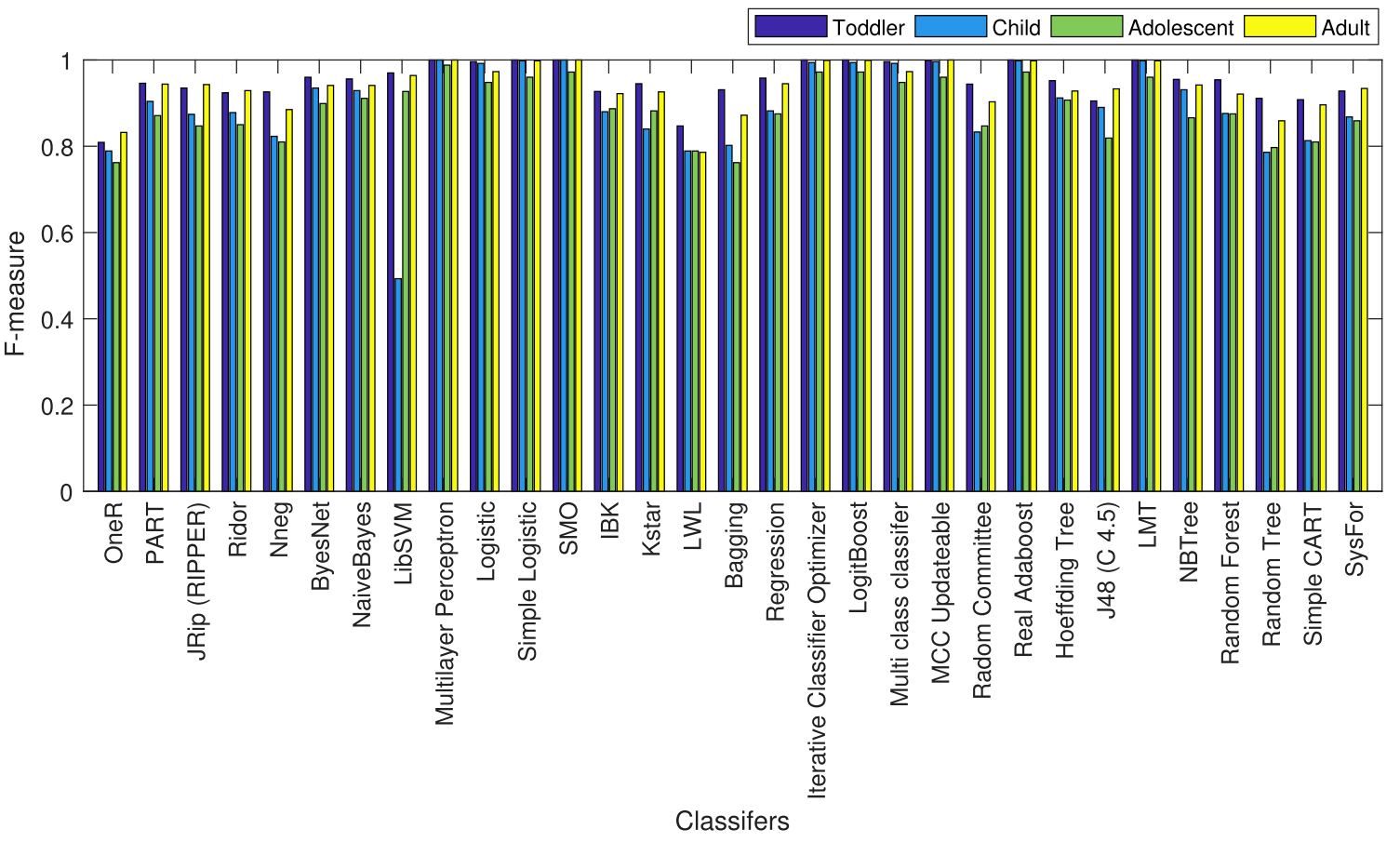}
    \caption{Comparison of classifier Performance (F-Measures)}
    \label{fig:ClassPerf}
\end{figure*}

We present the classification results in this section. Fig \ref{fig:ClassPerf} summarizes the results (sensitivity and specificity) obtained using classification analysis based on four ASD datasets. We find that mainly 4 out of 25 classifiers such as SMO, Logistic Regression, Multi Class Classifier and Multilayer perceptron (MLP) outperform others in terms of accuracy. Hence, we will be comparing and presenting results based on these algorithms henceforth. In our experiment, we aim to compare the classification errors of these four classifiers and identify the best performing classifier.

\subsection{Classification Errors}
Usually, we observe different types of error during the classification. We have different terminology to express these errors such as Deviation, Mean Absolute Error, and Root mean Square Error etc. These errors are the outcome of the differences of our predicted and observed data. We will determine which classifier has the least error. Before getting started, we learn about these different classification errors.

Mean Absolute Error (MAE):
MAE is a linear score which means that all individual differences are weighted equally in the average. It is the average over the test sample of the absolute differences between prediction and actual observation where all individual differences have equal weight.

\begin{equation}
    MAE= \frac{1}{n}\sum_{j=1}^{n}\left |Y _{j}-\bar{Y_{j}} \right |              
\end{equation}

Root Mean Square Error (RMSE): 
It is the square root of the average of squared differences between prediction and actual observation.

\begin{equation}
RMSE=\sqrt{\frac{\sum_{j=1}^{n}(Y_{j}-\bar{Y_{j}})^{2})}{n}}                    
\end{equation}

Relative absolute error (RAE):  
It is expressed as a ratio, comparing a mean error (residual) to errors produced by a trivial or naive model. A reasonable model (one which produces results that are better than a trivial model) will result in a ratio of less than one. The first version of Thiel’s U, called “U1” is a measure of forecast accuracy, comparing actual earnings (A) to Predicted earning (P).

\begin{equation}
RAE(U_{1})=\frac{\sqrt{\sum_{i=1}^{n}(P_{i}-A_{i}})^{2})}{\sqrt{\sum_{i=1}^{n}A_{i}^2}}                  
\end{equation}

Root relative squared error (RRSE): It is a simple predictor. It is just the average of the actual values. Thus, the relative squared error takes the total squared error and normalizes it by the total squared error of the simple predictor.

\begin{equation}
RRSE=\sqrt{\frac{(P_{1}-A_{1})^{2}+...(P_{n}-A_{n})^{2}}{(A_{1}-\bar {A})^{2}+...(A_{n}-\bar {A})^{2}}}               
\end{equation}%

\begin{table*}[htbp]
\centering
\caption{Classification Errors}
\resizebox{1\textwidth}{!}{
\begin{tabular}
{
m{3.1cm}
m{0.7cm}p{0.7cm}m{0.7cm}m{0.7cm}
m{0.05cm}
m{0.7cm}p{0.7cm}m{0.7cm}m{0.7cm}
m{0.05cm}
m{0.7cm}p{0.7cm}m{0.7cm}m{0.7cm}
m{0.05cm}
m{0.7cm}p{0.7cm}m{0.7cm}m{0.7cm}
}
\hline
\multirow{2}{*}{Classifier}  
&\multicolumn{4}{c}{Toddler} &
&\multicolumn{4}{c}{Child} &
&\multicolumn{4}{c}{Adolescent}&
&\multicolumn{4}{c}{Adult}\\ 

\cline{2-5}\cline{7-10}\cline{12-15}\cline{17-20}

& MAE & RMSE & RAE & RRSE &
& MAE & RMSE & RAE & RRSE &
& MAE & RMSE & RAE & RRSE &
& MAE & RMSE & RAE & RRSE \\ 
\hline
Multilayer~Perceptron  
& 0.10 & 0.23 & 0.23 & 0.50 &
& 0.20 & 0.56 & 0.40 & 1.11 &
& 1.98 & 10.54 & 3.95 & 21.08 &
& 0.10 & 0.30 & 0.22 & 0.63\\

SMO
&  0 &	0  & 0	& 0 &
&  0 &	0  & 0	& 0 &
& 2.82	& 16.8 & 5.648 & 33.61 &
&  0 &	0  & 0	& 0\\

Logistic
& 0.38	& 6.16 & 0.8878	& 13.33 &
& 1.15	& 10.28	& 2.302	& 20.56 &
& 5.42	& 22.25	& 10.847 & 44.51 &
& 2.61	& 16.01	& 6.0022 & 34.32\\

Multi Class Classifier
& 0.38	& 6.16	& 0.89 & 13.33 &
& 1.15	& 10.28	& 2.30	& 20.56 &
& 5.42	& 22.25	& 10.85 & 44.51 &
& 2.61	& 16.01	& 6.01 & 34.32\\

\hline

\hline
\multicolumn{17}{p{1\textwidth}}{}\\
\multicolumn{17}{p{1\textwidth}}{NOTE: `MAE' means Mean Absolute Error, `RMSE' means Root Mean Squared Error, `RAE' means Relative Absolute Error and `RRSE' means Root Relative Squared Error}
\end{tabular}
}
\label{tab:errors}
\end{table*}

From the results presented in Fig.~\ref{fig:ClassPerf} and Table~\ref{tab:errors}, we can see that the best performed classifier is SMO (Sequential minimal optimization). Though MLP has competitive performance but it has more classification errors and execution time requirements (~160-180\% higher than SMO). MLP classifier exhibits errors in all our ASD datasets, whereas SMO exhibits classifier's only for adolescent datasets. Sequential minimal optimization(SMO) implements John Platt's sequential minimal optimization algorithm for training a dataset. The SMO based classifier provided us the bets outcome among all other classifiers  our ASD datasets.  Besides, logistic regression and multi-class classifiers exhibit identical error rates and holds the third position together.

\section{Feature Engineering}
\label{sec:2}
Fadi et al. \cite{Thabtah2019Regree} adopt CHI and IG feature selection methods to seek similarities and differences in the feature sets offered by the methods. Authors also conclude that Logistic regression performs well in Adolescent and Adult dataset. In contrast, we perform our analysis on all four ASD datasets and our research concludes that Support Vector classifier- SMO perform well in all the four datasets.   Now, in order to find the best feature selection methods, we apply most popular five feature selections methods on our four ASD datasets and compare side by side (see Table~\ref{tab:fscomparison}).

\begin{table*}[htbp]
\centering
\caption{Comparison of feature selection methods}
\resizebox{1\textwidth}{!}{

\begin{tabular}{
m{0.6cm}
m{0.8cm}p{0.5cm}m{1.2cm}m{0.8cm}
m{0.05cm}
m{0.8cm}m{0.5cm}m{1.2cm}m{0.8cm}
m{0.05cm}
m{0.8cm}m{0.5cm}m{1.2cm}m{0.8cm}
m{0.05cm}
m{0.8cm}m{0.5cm}m{1.2cm}m{0.8cm}
m{0.05cm}
m{0.8cm}m{0.5cm}m{1.2cm}m{0.8cm}
}
\hline
\multirow{2}{*}{Rank} &\multicolumn{4}{c}{Information Gain} &
&\multicolumn{4}{c}{Chi Squared} &
&\multicolumn{4}{c}{Correlation}&
&\multicolumn{4}{c}{One R}&
&\multicolumn{4}{c}{Relief F}\\ 

\cline{2-5}\cline{7-10}\cline{12-15}\cline{17-20} \cline{22-25}

& Toddler & Child & Adolescent & Adult &
& Toddler & Child & Adolescent & Adult &
& Toddler & Child & Adolescent & Adult &
& Toddler & Child & Adolescent & Adult &
& Toddler & Child & Adolescent & Adult\\
\hline
\hline

1 
&	A9	& A4 & A6 & A6 &
& A9 & A4 & A6 & A6 &
& A9 & A4 & A6 &	A6 & 
&	A7 & A4 & A6 & A6 &
&	A9 & A4 & A6 & A5\\

2 
&	A5 & Res & Res & A5 &
& A6 & A6 & A3 & A9 &
& A6 & A6 & A3 & A9 &
& A6 & A9	& A3 & A9 & 
&	A5 & A8 & A3 & A6\\

3 
& A6 & A6 & A3 & A9 &
& A5 & A9 & Res &	A5 &	
& A5 &	A9 &	A4 &	A5 & 
&	A5 &	A8 &	A4 &	A5 & 
&	A2 &	A9 &	A5 &	A9\\

4 
& A7 & A9 & A4 & Res &
& A7 & Res & A4 & A4 &
& A7 & Res & A5 & A4 &
& A9 & A6 & A5 & A4 &
& A6 & A10 & A4 & A3\\

5 
& A4 & A8 & A5 & A4 &
& A4 & A8 & A5 & Res &
& A4 & A8 & A9 & A3 &
& A1 & A5 & A9 & A3 &
& A7 & A1 & A7 & A4\\

6 
& A1 & A5 & A9 & A3 &
& A1 & A5 & A9 & A3	&
& A1 & A5 & A10	& A10 & 
&	A4 & A10 & A10 & A7 & 
& A4 & A5 &	A9 & A10\\

7
& A2 & A3 & A10 & A10 &
& A2 & A3 & A10 & A10 &
& A2 & A3 & A7 & A7 &
& Age & A3 & A7 & Res & 
& A1 & A6 & A10 & A7\\

8 
& A8	& A10 & A7 & A7	&
& A8 & A10 & A7	& A7 &
& A8 & A10	& A2 &	A2 & 
& A3	& A1 & A2 & A2 &
& A8 & A3 & A2 &	A1\\

9 
& A3	& A1 & Ethn & Ethn &
& A3 & A1 & Ethn & Ethn &
& A3 & A1 & A1 &	A1 &
& FASD & A7 & A1 & A8 &
& A3 &	A7 & A1	& A8\\

10 
& Ethn & A7 & A2 & A1 &
& Ethn & A7 & A2 & A2 &
&	A10 & A7 & A8 &	A8 &
& Jaun & A2 & A8 & A1 &
& A10 & A2 & A8 & A2\\

11 
& A10 & A2	& A1 &	A2 &
& A10 & A2 & A1 &	A1 &
& Sex & A2 &	Ethn &	Ethn &
& A10 & Res & Res &	A10 &
& Ethn & Res & Ethn &	Res\\

12
& Age & Ethn & A8 &	A8 &
& Age & Ethn & A8 & A8 &
& Jand & Ethn &	Res & FASD &
& A8	& Ethn &	Ethn & Jand &
& Jand & Age	& Res &	Ethn\\

13 
& Sex & Sex & Jand &	FASD & 
& Sex & Sex & Jand & FASD &
& Ethn & Sex & Age & Res &
& Sex &	Age & Age & Sex &
& Age &	Sex & Age & Jand\\

14 
& Jand & FASD & Sex	& Jand &
& Jand &	FASD & Sex & Jand &
& Age & FASD & Jand &	Jand &
&	A2 & Sex & Jand & Ethn &
& FASD & Jand &	FASD & Age\\

15 
& FASD & Jand & FASD	& Sex &
& FASD &	Jand &	FASD & Sex &
& FASD & Jand & Sex	& Age &
&	Ethn & Jand & Sex & FASD &
& Sex &	FASD &	Sex &	Sex\\

16 
& -- & Age & Age & Age &	
& -- &	Age &	Age &	Age &
& -- & Age & FASD & Sex &
& -- & FASD & FASD	& Age& 
& -- &	Ethn & Jand & FASD\\
\hline

\hline
\multicolumn{21}{p{.9\textwidth}}{}\\
\multicolumn{21}{p{.9\textwidth}}{NOTE: `Res' means `Residence', 'Ethn' means 'Ethnicity', `FASD' means 'Family ASD', `Jand' means Jaundice}
\end{tabular}
}
\label{tab:fscomparison}
\end{table*}

\subsection{Analyzing feature selection techniques and feature ranking}
In this section, we explore the effectiveness of feature selection on our ASD datasets. We know that the attributes which are the answers of the question Q1 to Q10, mainly the deciding factors of the ASD cases. Besides, answers of the demographic questions have little to no effect for identifying ASD cases. So, by comparing the five feature selection methods, we find that Relief F attribute selection method performs best among the five methods and it is capable to rank the AQ attributes before the demographic attributes for all the four ASD datasets.

\begin{figure}
    \includegraphics[width=1\columnwidth]{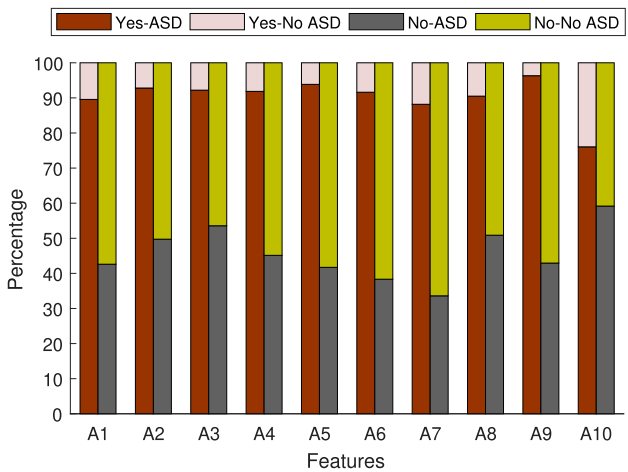}
    \caption{AQ-10 questions responses for Toddler ASD dataset}
    \label{fig:Attr_todd}
\end{figure}

\begin{figure}
    \includegraphics[width=1\columnwidth]{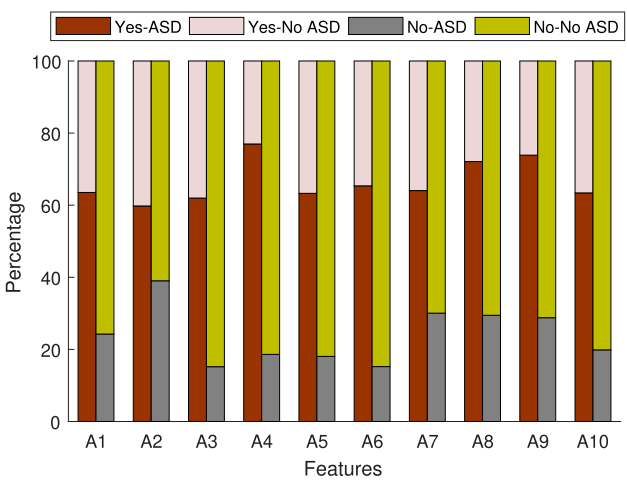}
    \caption{AQ-10 questions responses for Child ASD dataset}
    \label{fig:Attr_child}
\end{figure}

\begin{figure}
    \includegraphics[width=1\columnwidth]{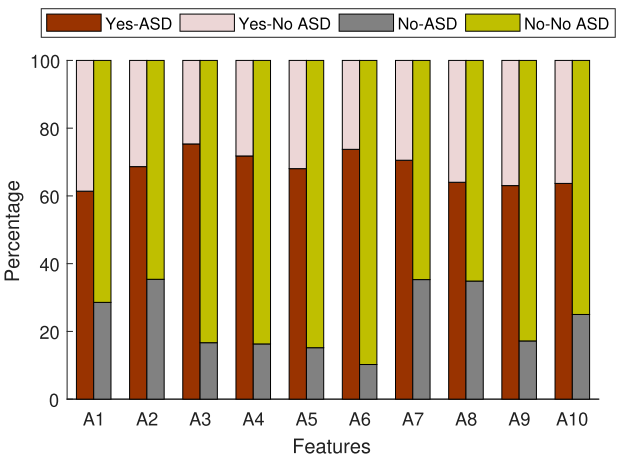}
    \caption{AQ-10 questions responses for Adolescent ASD dataset}
    \label{fig:Attr_adoles}
\end{figure}

\begin{figure}
    \includegraphics[width=1\columnwidth]{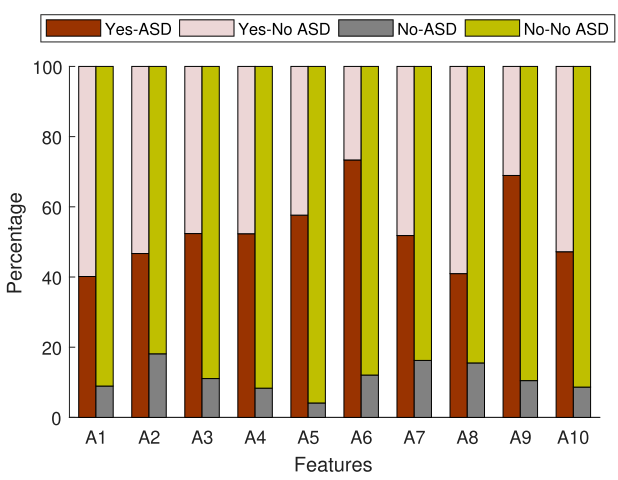}
    \caption{AQ-10 questions responses for Adult ASD dataset}
    \label{fig:Attr_Adult}
\end{figure}

\begin{figure}
    \includegraphics[width=1\columnwidth]{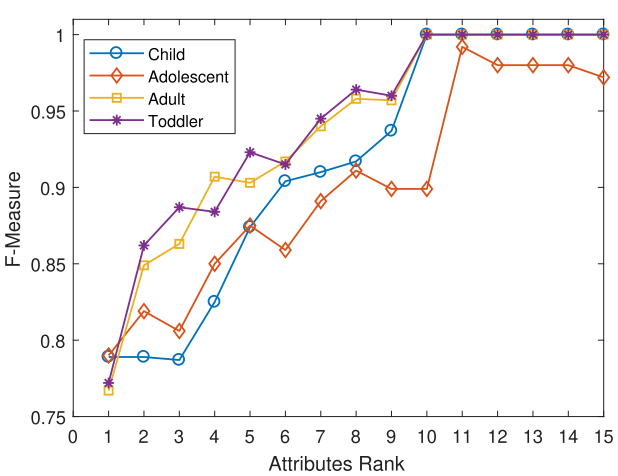}
    \caption{SMO Classifier’s Performance over increasing attributes for four ASD datasets}
    \label{fig:SMO Performance}
\end{figure}

We count the total number of occurrences of the attributes in four ASD datasets respectively and compare the effects of the attributes in detecting ASD cases. In Fig. 7-10, the first column in each group represent the score `1' of the attribute when the ASD case is "yes" (lower part) and "no" is the upper part of the column and the second column represent the score `0'  of the attribute when the ASD case is "yes" represent the lower part and ``no'' represent the upper part.We compare if the attribute has a better ratio in detecting ASD ``yes'' and “no” cases in the first and second column respectively. Thus if we rank these attributes, we find that it supports Relief F feature selection method where the rank remain same as per individual attribute analysis for each age groups.

By observing the toddler dataset, we find the top three attributes are A9, A5 and A2; for child, these are A4, A8 and A9; for adolescent, these are A6, A3 and A5 for Adult, these are A5, A6 and A9.

We take the consideration of Relief F attribute selection method and compare the feature performance using our selected classifiers- SMO by plotting F-measure value against the increment of attribute number.

\subsection{Identifying most significant features}

We analyze the classifier performance by increasing the attribute numbers and compare the F-measure values against this. We apply this for both SMO and Multilayer Perceptron (MLP).
We find that the F-measures value increases according to the increment of the attribute numbers. It reaches the highest point (e.g ‘1’) when the total number of attributes is 10, as shown in \ref{fig:SMO Performance}. After that it becomes straight line for Toddler, Child and Adult datasets but for Adolescent dataset it tends to 0.972 when the attribute number count upto 10 and then fall little bit with the increment of attribute numbers. 
\begin{table*}[htbp]
    \centering
    \caption{Comparison of our approach with the state-of-the-art approach}
    \begin{tabular}{c cc cc cc cc}
    \hline
    \multirow{2}{*}{Approach}&
    \multicolumn{2}{c}{Toddler}&
    \multicolumn{2}{c}{Child}&
    \multicolumn{2}{c}{Adolescent}&
    \multicolumn{2}{c}{Adult}\\
     & Recall & Accuracy &
     Recall & Accuracy & 
     Recall & Accuracy & 
     Recall & Accuracy\\
    \hline
    \hline
    
    Fadi et al.~\cite{Thabtah2019Regree, thabtah2020new} &-&-&0.91&91& 0.99 & 99.91 & 0.97 & 97.58\\
    K. Basu et al.~\cite{kbasu}
    &-&-&-&-&-&-& 1 & 100\\
    Brian McNamara et al.\cite{BrainMc}
    &-&-&-&-&-&-& 0.93 & 91.74\\
    Our approach &1&100&1&100& 0.99 & 99.19 & 1.0 & 100\\
    
    \hline
    
    \hline
    \end{tabular}
    \label{tab:perfcomp}
\end{table*}
\section{Classification Results Comparison}

In this section, we compare the performance of our proposed method with the existing benchmark algorithms \cite{Thabtah2019Regree, thabtah2020new}. In~\cite{Thabtah2019Regree}, authors use a logistic regression based framework to reveal important information related to autism screening. Authors consider only IG and Chi squared feature selection method. In \cite{thabtah2020new}, authors utilizes a Rules-Machine Learning for the same purpose. In those works, authors analyze the results using Child, Adult and Adolescent datasets. For those datasets, authors obtain 99.9\% and 97.58\% accuracy for Adolescent and Adult datasets, respectively as presented in Table~\ref{tab:perfcomp}. However, the accuracy from the Child dataset is only 91\%. Compared to those, our proposed method has significantly higher accuracy as shown in the table. The accuracy and recall obtained from Basu et al.~\cite{kbasu} is very high and similar to our proposed method, however, authors have demonstrated the utility only for one dataset. Similarly, Brian et al. in their work \cite{BrainMc} obtained 91.74\% accuracy and 93\% recall using the same Adult dataset. However, both of the aforementioned research considered only Adult dataset. We have evaluate our proposed method considering all four datasets and results show that our proposed method outperforms other datasets in terms of both accuracy and recall for all four experimental setups.



\section{Conclusion}
In this study, we have analyzed the ASD datasets for Toddler, Child, Adolescent and Adult. After applying different classification techniques, we conclude that SMO (Sequential minimal optimization) based Support Vector Machine (SVM) algorithm works best for the detection of ASD cases for all ASD datasets. We use 10-fold cross validation method to evaluate our dataset. Besides, we apply Attribute selection methods to derive fewer features from ASD screening methods yet maintain competitive performance. We have applied five attribute section methods on ASD datasets. Our finding also reveals that Relief F feature selection methods outperform amongst others. The main limitation of this research is the restricted access and lack of enough ASD datasets. In our future work, we plan to work with deep learning methods that integrate feature assessment and classification altogether for improved performance.

\bibliographystyle{spphys}       
\bibliography{MyCollection}   

\end{document}